\documentclass[11pt,a4paper]{article}
\usepackage[hyperref]{ranlp2023}
\usepackage{times}
\usepackage{latexsym}
\usepackage{graphicx}

\usepackage{microtype}

\aclfinalcopy 

\title{Automatic Extraction of the Romanian Academic Word List: Data and Methods}

\author{Ana-Maria Bucur$^{1,2}$, Andreea Dincă$^2$, Mădălina Chitez$^2$, Roxana Rogobete$^2$\\
$^1$ Interdisciplinary School of Doctoral Studies, University of Bucharest\\
$^2$ West University of Timișoara, Romania\\
ana-maria.bucur@drd.unibuc.ro\\
\{madalina.chitez, andreea.dinca, roxana.rogobete\}@e-uvt.ro\\
}

\date{}

\begin{document}
\maketitle
\begin{abstract}
This paper presents the methodology and data used for the automatic extraction of the Romanian Academic Word List (Ro-AWL). Academic Word Lists are useful in both L2 and L1 teaching contexts. For the Romanian language, no such resource exists so far. Ro-AWL has been generated by combining methods from corpus and computational linguistics with L2 academic writing approaches. We use two types of data: (a) existing data, such as the Romanian Frequency List based on the ROMBAC corpus, and (b) self-compiled data, such as the expert academic writing corpus EXPRES. For constructing the academic word list, we follow the methodology for building the Academic Vocabulary List for the English language. The distribution of Ro-AWL features (general distribution, POS distribution) into four disciplinary datasets is in line with previous research. Ro-AWL is freely available and can be used for teaching, research and NLP applications.
\end{abstract}

\section{Introduction}
Since academic language differs from everyday social language and is an essential acquisition target in current education, extracting salient features contributes to linguistic, register, genre and disciplinary feature identification that can benefit students, teachers and educational app developers alike. Compiling an Academic Word List (AWL) is an effective solution to support both language teaching and NLP tasks. From the didactic perspective, AWLs reflecting either the L1 (i.e. mother tongue) or the L2 (i.e. foreign language) academic vocabulary can be used to offer linguistic support to novice academic writers in the form of discipline-specific and general lexical prompts. Teachers of all disciplines can integrate AWLs into teaching materials to help students write better (see, for example, \citet{wangdi2022investigating}).

NLP studies can exploit AWL datasets on topics such as text classification \cite{zampieri2012evaluating} and topic modelling \cite{murakami2017corpus}. For example, field-specific academic lists can be used to automatically classify texts into disciplinary areas. The same can be applied for the automatic distribution of texts in academic versus non-academic batches. In machine learning methods for language modelling tasks, AWLs are essential in training models to generate accurate academic writing samples. By combining NLP tasks with linguistic approaches in relation to AWLs, important advances can be achieved in the frame of lexical and syntactic analyses that evaluate the use of collocations and phraseology specific to the academic varieties. For the Romanian language, there have been few attempts to extract a valid Romanian Word List \cite{szabo2015introducing} and only one study has extracted and analysed multiword units using academic writing corpora \cite{muresan2022phraseology}. 

In recent years, researchers have worked to create several academic writing corpora. EXPRES – Corpus of Expert Writing in Romanian and English \cite{chitez2022expres} is one of them. It is the only bilingual multidisciplinary corpus capturing the Romanian academic writing context. By combining datasets representing the Romanian Frequency List \cite{szabo2015introducing} based on the ROMBAC Corpus \cite{ion2012rombac}, and EXPRES disciplinary datasets \cite{chitez2022expres}, we were able to generate an empirically based Romanian Academic Word List. Ro-AWL is made publicly available\footnote{\url{https://github.com/bucuram/Ro-AWL}} and can be used for teaching, text classification and language modelling.

\section{Related Work}
Most academic vocabulary lists have been developed in the context of English for Academic Purposes (EAP). On the whole, two categories of lists exist. One list type aims to identify academic words commonly used in EAP across disciplines, which students could be made aware of. The studies aiming to provide cross-disciplinary academic word lists usually use large corpora containing expert academic writing from various disciplines. The widely used lists of this type are the Academic Word List (AWL) \cite{coxhead2000new} and the Academic Vocabulary List (AVL) \cite{gardner2014new}. The second type of list seeks to identify discipline or field-specific words worth teaching. Various specialised lists have been developed for fields such as veterinary medicine \cite{ohashi2020esp} or nursing \cite{yang2015nursing}.

While there is a growing interest in building cross-disciplinary academic word lists for languages other than English, these academic word lists remain few. See, for example studies conducted for French \cite{cobb2004there}, Persian \cite{rezvanifirst}, Portuguese \cite{baptista2010p}, Swedish \cite{carlund2012academic}, and Norwegian \cite{johannessen2016constructing}. An explanation for this scarcity might be that academic language data sets are rare and often not freely available due to copyright. This can be especially true for low-resource languages, such as Romanian. Access to a representative corpus is crucial, as the validity and reliability of an academic word list highly depend on the quality of the data set. 

Apart from the limited availability of academic writing corpora, an additional challenge may be that there is no standard procedure for extracting academic word lists. Scholars are still exploring and testing various methodologies. For example, some studies build on the methods used for the AWL or the AVL \cite{johannessen2016constructing,rezvanifirst}. One study uses the translated version of the AVL in Portuguese as a starting point for its investigation \cite{baptista2010p}. Another study proposes a new word list extraction method different from previous ones \cite{carlund2012academic}.  

In the case of Romanian, no previous studies have compiled specialised or general academic word lists. Although in the last 10-15 years, several research institutions and projects have been involved in developing corpus resources in Romanian, relatively few have focused exclusively on general academic writing. Some of the most significant corpora recently compiled, such as ROMBAC (Romanian Balanced Annotated Corpus, see \citet{ion2012rombac}), with more than 30 million words, CoRoLa (Corpus of Contemporary Romanian Language, see \citet{mititelu2014corola}), or The Balanced Romanian Corpus (BRC, see \citet{midrigan2020resources}) cover only few disciplines or subsets: 5 sections for ROMBAC (journalism, literature, medical texts, legal texts, biographies), uneven and unfiltered distribution of resources in CoRoLa (the collection of academic writing texts is based on agreements with publishing houses and journals, without filtering of the content on quality criteria) and BRC (literary text samples, research articles, news, spoken data). The ROMBAC corpus (excluding the medical subcorpus) was already used to develop the Romanian Word List (RWL, see \citet{szabo2015introducing}), targeted at Romanian L2 learners (e.g. from the Hungarian minority in Romania). The list is a general list of words, not focused on academic language. As far as discipline-specific corpora are concerned, smaller corpora such as SiMoNERo (medical corpus, \citet{mitrofan2019monero}), BioRo \cite{mitrofan2018bioro}, PARSEME-Ro (news articles), LegalNERo (legal, \citet{paiș2021named}), MARCELL (legal, multilingual, see \citet{varadi2020marcell}), CURLICAT (multilingual, containing several domains: Economics, Education, Health, Sciences, etc., see \citet{varadi2022introducing}) have been compiled. However, apart from compiling the datasets and conducting a series of descriptive studies, no special attention is given to the lexical level. 

In this context, the EXPRES corpus (Corpus of Expert Writing in Romanian and English) is the first corpus of discipline-specific academic writing in the Romanian context (academic writing in Romanian L1 and academic writing in English L2 produced by Romanians) \cite{bucur2022expres,chitez2022write}. Covering four disciplines – Linguistics, Economics, Political Sciences, Information Technology –, the Romanian subset contains 200 open-access research articles from each domain, published in the past 5-10 years in peer-reviewed journals (see \citet{chitez2022expres}). The rigorous selection criteria \cite{rogobete2021challenges} contribute to the representativeness of the corpus, making it a suitable candidate for testing a possible Romanian Word List and narrowing it down to an Academic Word List. Furthermore, the EXPRES corpus is the first Romanian expert academic corpus available on an open-access query platform. Unlike other Romanian corpora, which offer limited access to third parties and poor resources for downloading search results or statistics, the EXPRES corpus support platform has been implemented as a cross-platform distributed web application  \cite{chitez2022expres}.

\section{Data}
This work uses two different corpora: the academic corpus EXPRES and the Romanian Academic Word List \cite{szabo2015introducing} compiled from the general corpus ROMBAC. The Romanian language sub-corpus of EXPRES\footnote{\url{https://expres-corpus.org/}} \cite{chitez2022expres} consists of 800 research articles, 200 articles for each of the four fields: Linguistics (LG), Economics (EC), Information Technology (IT) and Political Sciences (PS). The articles from the corpus were manually processed to preserve the anonymity of the authors (e.g., the name of the authors were replaced with {AUTHOR\_NAME}) and the beginning and end of the title, abstract and sections are annotated with corresponding XML tags (e.g., $<$TITLE$>$, $<$/TITLE$>$) \cite{chitez2022expres}. Table \ref{tab:expres} shows the distribution of words in EXPRES, without counting the manually added tags. The corpus contains more than 3 million words, with more than 200 thousand unique words.

\begin{table}[hbt!]
    \centering
    \caption{EXPRES Statistics}
    \resizebox{0.6\linewidth}{!}{
    \begin{tabular}{l|ll}
    \textbf{Domain}  & \textbf{Tokens} & \textbf{Types}\\
    \hline
    \textbf{EC} & 1,092,846 & 48,807 \\
    \textbf{LG} & 674,277 & 73,667 \\
    \textbf{IT} & 750,236 & 40,494 \\
    \textbf{PS} & 963,061 & 62,096 \\
    \hline
    \textbf{Total} & 3,480,420 & 225,064\\
    \end{tabular}
    }
    \label{tab:expres}
\end{table}

The Romanian Academic Word List \cite{szabo2015introducing} contains a frequency list for all the words in the Romanian Balanced Annotated Corpus (ROMBAC) \cite{ion2012rombac}. ROMBAC \cite{ion2012rombac} is a large general collection of texts from the Romanian language. It contains texts from five domains: news, medical, legal, biographies and fiction. The texts from ROMBAC are tokenized and lemmatized. The version we use in this paper contains more than 25 million lemmas, of which 1 million are unique (Table \ref{tab:rombac}). The dataset was previously used to derive other linguistic resources, such as the Romanian Word List and Romanian Vocabulary Levels Test \cite{szabo2015introducing}. We use the ROMBAC corpus in our work because it is the largest corpus available in Romanian that was not web-scraped, and it is a reference corpus for the contemporary Romanian language \cite{ion2012rombac}. Even if another larger corpus for the contemporary Romanian language exists, namely CoRoLa \cite{mititelu2014corola}, it is not publicly available and cannot be downloaded; it can only be queried online\footnote{\url{https://korap.racai.ro/}}. The other reference corpus recently compiled, BRC \cite{midrigan2020resources}, was not an option either, since its size is smaller than ROMBAC and lacks disciplinary variation.

\begin{table}[hbt!]
    \centering
    \caption{ROMBAC Statistics}
    \resizebox{0.7\linewidth}{!}{
    \begin{tabular}{l|ll}
    \textbf{Domain}  & \textbf{Tokens} & \textbf{Types}\\
    \hline
    \textbf{News} & 1,922,109 & 50,945 \\
    \textbf{Medical} & 6,783,005 & 362,782 \\
    \textbf{Legal} & 6,269,543 & 248,354 \\
    \textbf{Biographies} & 3,716,031 & 223,592 \\
    \textbf{Fiction} & 6,950,371 & 105,346 \\ 
    \hline
    \textbf{Total} & 25,641,059 & 991,019 \\ 
    \end{tabular}
    }
    \label{tab:rombac}
\end{table}

\section{Methodology}
\textbf{Data preprocessing.} The Romanian Academic Word List, with words from the ROMBAC corpus, provides the lowercase lemma for each word in the corpus and its frequency. Therefore, no preprocessing step was done on this data. Even if we use the word frequencies from the Romanian Academic Word List, we will refer to this data as the ROMBAC corpus, given that the list contains all the words from ROMBAC.

The EXPRES corpus is organised in multiple .txt files, one for each article from the four domains LG, IT, PS, and EC. For each document, we removed specific tags used for article anonymisation, such as {JOURNAL\_TITLE}, {AUTHOR\_NAME}, etc., and the specific XML tags used to mark the beginning or end of the title ($<$TITLE$>$, $<$/TITLE$>$), abstract ($<$ABS\_INT$>$, $<$/ABS\_INT$>$), or different sections of the article ($<$INTROD$>$, $<$/INTROD$>$), etc. The EXPRES corpus statistics regarding the words and word types in the corpora are shown in Table \ref{tab:expres}. For preprocessing the text, we used Stanza \cite{qi2020stanza} for lemmatising and extracting part-of-speech tags. All the lemmas from the texts are transformed into lowercase. The Stanza toolkit was chosen for its good performance for the Romanian language, compared to other NLP tools \cite{pais2021depth}. However, we performed a manual analysis of the extracted lemmas and observed that some of them are incorrect: “sociales” instead of “social” ( En: “social”), “europes” instead of “european” (En: “European”), and others. Even if previous works have shown a good performance of the Stanza toolkit for lemmatisation in the Romanian language \cite{pais2021depth}, we chose to use the lemmas from the ROMBAC corpora for the words that appear in ROMBAC. We used Stanza only for extracting the lemma of words that were not part of ROMBAC. This way, the noise of lemmatisation was diminished, as the lemmas provided in the ROMBAC corpus were accurate and have been previously validated \cite{ion2012rombac}. 

\textbf{Building the academic word list.} For constructing the academic word list, we follow the methodology for building the Academic Vocabulary List for the English language \cite{gardner2014new}, comprising different frequency measures for lemmas. We chose to use the methodology from \citet{gardner2014new} instead of the procedure from \citet{coxhead2000new} because the former method provides an academic list with almost twice the latter's coverage. The approach from \citet{coxhead2000new} is based on word families, while the method from \citet{gardner2014new} relies on lemmas. A word family is represented by the base word from which other words are derived with suffixes and prefixes. This can be problematic in the case of academic words, as the base of a word family can be an academic word, but their derivations might not be academic \cite{gardner2014new}.

The methodology is based on four measures: ratio, range, dispersion and discipline measure. The ratio is used to exclude general high-frequency words from the corpus, while the other three metrics exclude technical or discipline-specific terms. We further expand on each metric below.

\textbf{Ratio.} Similar to \citet{gardner2014new}, general high-frequency words (in our case, lemmas) are removed from the academic word list. The ratio is computed to keep in the list words with a higher frequency in the academic corpus than in the general non-academic corpus. We computed the normalised frequency per million words of each word in the two corpora, EXPRES and ROMBAC. The ratio is calculated by dividing the academic corpus's normalised frequency by the general corpus's normalised frequency for each word. \citet{gardner2014new} use the frequency ratio of 1.5 in their method, but mention that the measure is not a gold standard. We experimented with values between 1.2 and 2.0 for ratio, and, in our case, the 1.2 ratio was a suitable value, to not have important academic words excluded from our list, such as “metodologic" (En: “methodological"), “clasificare" (En: “classification"), “activitate" (En: “activity"), “distinge" (En: “distinguish"), “sugera" (En: “suggest"), which are found in the original AVL for the English language.

\textbf{Range.} The range measure allows for selecting words that only occur in multiple disciplines, and filtering out discipline-specific words. \citet{gardner2014new} proposed that a word should have at least 20\% of the expected frequency in 78\% of the sub-corpora (i.e. 7 out of 9 domains). For computing the expected frequency, we first calculated each word’s frequency in relation to the corpus by dividing the word count by the total number of words in EXPRES. Afterwards, the frequency in relation to the corpus is multiplied by the number of words in a given sub-corpora to get the expected frequency in each sub-corpora.
In our case, EXPRES has only four domains, and we chose words that had at least 20\% of the expected frequency in at least three out of four fields, corresponding to 75\% of sub-corpora.

\textbf{Dispersion.} The measure used for dispersion is Julliand's D \cite{juilland2021frequency}, which shows how evenly a word appears in a corpus. The formula is as follows:

\vspace{3mm}
$Juilland's D = 1 - \frac{\sigma }{\bar{x}}\times \frac{1}{\sqrt{n-1}}$
\vspace{3mm}

where $\sigma$ represents the standard deviation and $\bar{x}$ represents the mean frequency of a word. $n$ is the number of sub-corpora.

The values of dispersion range from 0.01 (corresponding to words that appear in a small part of the corpus) to 1.00 (meaning that a word is spread evenly in the corpus). Unlike the range measure, which estimates if a word has the expected frequency in the four domains, the dispersion measure ensures that a given word is distributed uniformly in the four sub-corpora. \citet{gardner2014new} chose 0.80 dispersion, while, in other works, the dispersion measure varies between 0.30 to 0.60 \cite{oakes2006use,johannessen2016constructing,lei2016new}. We decided to use a dispersion value of 0.50 in our work.

\textbf{Discipline measure.} This measure is used for filtering out words with a very high frequency in a given domain, which may be technical discipline-specific words. \citet{gardner2014new} proposed that a word cannot have more than three times the expected frequency in any domain. Following a similar approach, we remove words with more than three times the expected frequency in any of the four domains.

As an additional measure, we excluded words with low frequency in the academic corpus, because the metrics mentioned above do not filter them out. Inspired by \citet{coxhead2000new} and \citet{lei2016new}, we remove from the final academic list the words that have a minimum frequency of 28.57 per million words, corresponding to the minimum frequency originally proposed by \citet{coxhead2000new} of 100 times in the 3.5 million words corpus they used in their work. We also performed a manual analysis of the academic word list and removed the noise, such as proper nouns (e.g., “București”, En: “Bucharest”), some numerals, and some words that were not academic and that were not filtered out by the measures mentioned above.

\section{Results}
The final Romanian academic word list consists of 673 lemmas with their corresponding part of speech tags. The list comprises 332 nouns, 167 adjectives, 157 verbs, and 17 adverbs. We automatically translated into Romanian the words from the AVL developed for English \cite{gardner2014new} that contains 3015 words. We found that 381 words in our list are in the original AVL. There are some cases of academic words found in our Romanian academic list and in the AVL for English for which the automatic translation fails to provide the correct match. For example, the noun “adoption" from AVL was translated as “adopție", which is not in the Ro-AWL, but the word “adoptare" is an academic word from Ro-AWL with the same meaning. The fact that we found more than half of the Ro-AWL in the original AVL, even though in some cases the translation fails to capture the correct meaning of the words, makes us confident that the measures used are reliable for building a Romanian academic word list. 

\begin{figure}[!hbt]
    \centering
    \includegraphics[width=1\linewidth]{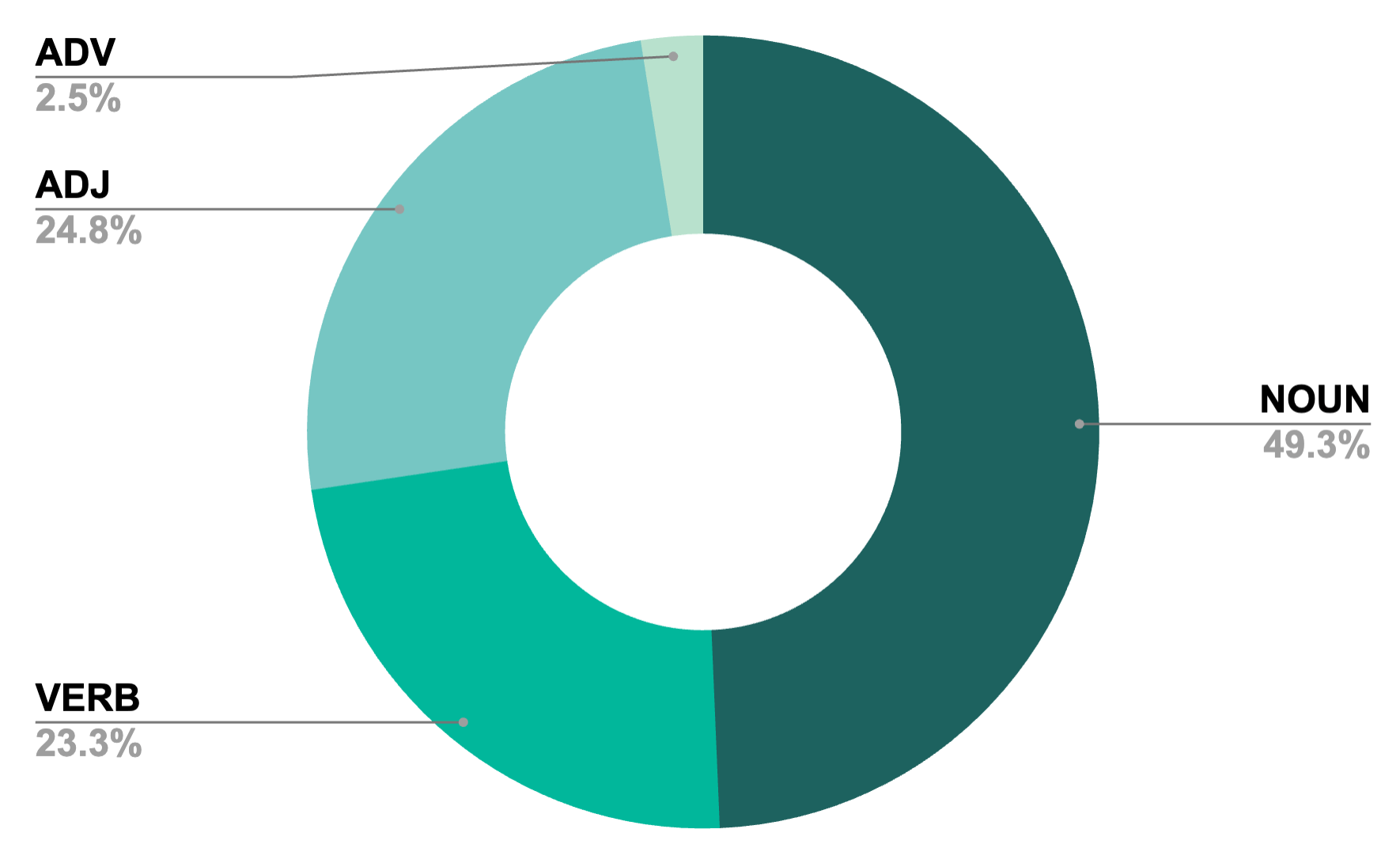}
    \caption{The distribution of the words in terms of part-of-speech from Ro-AWL}
    \label{fig:pos}
\end{figure}

In line with previous works \cite{gardner2014new,coxhead2000new,carlund2012academic}, to demonstrate the viability of the newly developed academic word list, we measured the coverage of the Ro-AWL in two corpora: the academic corpus EXPRES and in the general corpora ROMBAC. The academic words from our list cover 15.25\% of the EXPRES corpus and 6.73\% of ROMBAC. In line with the English AVL results, Ro-AWL has a higher coverage in the academic corpus and a lower coverage in the general corpus. Regarding the coverage in EXPRES, we show the coverage of academic words categorised by their part-of-speech tags in Table \ref{tab:coverage}. The coverage of the Romanian academic word list varies in the four domains. The coverage is 17.75\% for the Economics sub-corpora, 11.82\% for Linguistics, 17.03\% for Information Technology and 13.17\% for Political Sciences.
\begin{table}[hbt!]
    \centering
    \caption{Coverage of Ro-AWL in the EXPRES corpus}
    \resizebox{0.9\linewidth}{!}{
    \begin{tabular}{l|llll}
    \textbf{}  & \textbf{EC} & \textbf{LG} & \textbf{IT} & \textbf{PS}\\
    \hline
    VERB & 4.98\% & 3.95\% & 5.33\% & 3.95\% \\
    NOUN & 9.74\% & 6.02\% & 9.20\% & 6.82\% \\
    ADJ & 0.33\% & 0.27\% & 0.24\% & 0.16\% \\
    ADV & 2.70\% & 1.59\% & 2.26\% & 2.24\% \\
    \hline
    Total & 17.75\% & 11.82\% & 17.03\% & 13.17\% \\ 
    \end{tabular}
    }
    \label{tab:coverage}
\end{table}

\section{Discussion}
A first observation concerns the different coverages of Ro-AWL in the EXPRES corpus (see Table \ref{tab:coverage}). The lower percentages in Linguistics and Political Sciences (with a total coverage ranging between 11\% and 14\%) and the higher ones in Economics and IT confirm that “The SSH community is characterised by the embedment of research in the local context and by linguistic diversity in producing and disseminating knowledge” \cite{kancewicz2020does}. Researchers in the Romanian context in SSH (Social Sciences and Humanities) tend to favour a more “creative” dimension of the language used in academic writing, using figurative language in constructing rhetorical structures. Although in English language academic writing “the dichotomy of soft and hard sciences is rather fluid and as such insignificant” \cite{stotesbury2003evaluation}, discipline-specific peer-review practice in the Romanian setting seems to influence the academic writing style. This is particularly visible in the EXPRES subset of Political Sciences and Linguistics. Romanian academic writing in SSH seems rather unfocused, descriptive and rich in rhetorical structures.
In contrast, research articles in Economics and Information Technology contain many statistics, tables, and formulas, making the writing in the discipline less descriptive.

Secondly, although our extraction measures were successful in filtering most of the technical vocabulary, small amount of technical language remains in the Ro-AWL (terms such as “dauna”, En: “damage” - in contexts related to insurances; “institutional”, “security”, “electronic” etc.). Nevertheless, the majority of the Ro-AWL components are discipline neutral, thus contributing to academic discourse cohesion and coherence.

Thirdly, a technical challenge regarding the functionality and accuracy of the Romanian POS tagger should be mentioned. An overview of the assigned tags revealed the difficulty of the tagger to distinguish between adjectives and adverbs (for instance: “important”, “social”, “european” were assigned as adverbs, but the contexts prove their prevalent use as adjectives). It also confused past participles ending with “-t” (e.g. “accentuat”, En: “emphasised”. This technical difficulty can be observed in Table \ref{tab:coverage}, with the coverage of adverbs being higher than the one of adjectives, because most of the adjectives had the part-of-speech mislabeled by the POS tagger. These errors of the POS tagger are due to the homonymy between the two POS, most adverbs being homonymous to their adjective counterparts \cite{vasile2017properties}.

A technical advantage of the Romanian POS tagger, however, is its capacity to recognise nouns with a definite article while being a part of prepositional phrases (“în pofida”, En: “despite”, “în jurul”, En: “around”). This also explains the increased percentage levels of nouns, adverbs and verbs and the lower values for adjectives (see Figure \ref{fig:pos}). 

Despite some of the technical challenges, the extraction of the Romanian AWL using the EXPRES corpus resulted in successfully identifying the recurrent discourse conventions used by Romanian researchers. During the process and alongside the extraction procedure per se, translating the Academic Vocabulary List (AVL) \cite{gardner2014new} was a helpful procedure, as it is well accepted that academic writing, irrespective of the language, contains a large number of words of Greek and Latin origin (see e.g., \citet{rasinski2008greek,green2020greek}).

\section{Conclusions and Future Work}
This study reports the extraction of the first Romanian Academic Word List (Ro-AWL), which can be used to check the degree of academic vocabulary coverage in discipline-specific and general language samples. Ro-AWL consists of 673 lemmas, distributed among the main part-of-speech categories (nouns, verbs, adverbs, adjectives). Our methodology adopted measures used for the Academic Vocabulary List for the English language, such as ratio, range, dispersion and discipline measures. The percentages calculated by testing Ro-AWL on the disciplinary datasets in the EXPRES corpus \cite{chitez2022expres}, indicate a lower coverage for Linguistics (11.82\%) and Political Sciences (13.17\%) and a higher coverage for Information Technology (17.03\%) and Economics (17.75\%). Also, the academic vocabulary coverage in ROMBAC, a general language reference corpus, is 6.73\%, while the coverage is much higher (15.25\%) in EXPRES, an expert academic writing corpus. This aligns with previous research, since Ro-AWL coverage is similar to thresholds for academic vocabulary \cite{nation2001learning}.

Despite several computation constraints (e.g. Romanian POS tagger not being able to distinguish between adjectives and adverbs), our study provides important insights into the academic writing vocabulary in Romanian by proposing a validated Romanian Academic Word List. Our findings also have pedagogical implications, as the list can be used to support academic writing teaching activities and NLP tasks focusing on Romanian. For example, the Ro-AWL can be paired up with the freely available EXPRES corpus platform to develop corpus-assisted learning activities commonly known as Data-Driven Learning (DDL) (see e.g., \citet{bennett2010using}). However, even if the coverage test results in the EXPRES are encouraging, further research is needed to test the validity of the Ro-AWL on corpora containing academic writing from more disciplines. Future work can be conducted in at least two directions: refining the lists from a contrastive perspective, by developing a discipline-specific AWL, or, on the contrary, by searching for highly frequent academic words present in an extended corpus containing more disciplines.

\section*{Acknowledgements}
We would like to thank Csaba Z Szabo for giving us access to the Romanian Frequency List and Ștefan Daniel Dumitrescu for his valuable insights into the data and methods used in this paper. This work was supported by a grant of the Ministry of Research, Innovation and Digitization, CNCS/CCCDI – UEFISCDI, project number 158/2021, within PNCDI III, awarded to Dr Habil Madalina Chitez (PI), from the West University of Timisoara, Romania, for the project DACRE (\textit{Discipline-specific expert academic writing in Romanian and English: corpus-based contrastive analysis models}, 2021-2022).

\bibliography{refs}
\bibliographystyle{acl_natbib}

\end{document}